# VHPOP: Versatile Heuristic Partial Order Planner

**Håkan L. S. Younes**                                    LORENS@CS.CMU.EDU
**Reid G. Simmons**                                        REIDS@CS.CMU.EDU
*School of Computer Science, Carnegie Mellon University*
*Pittsburgh, PA 15213, USA*

## Abstract

VHPOP is a partial order causal link (POCL) planner loosely based on UCPOP. It draws from the experience gained in the early to mid 1990's on flaw selection strategies for POCL planning, and combines this with more recent developments in the field of domain independent planning such as distance based heuristics and reachability analysis. We present an adaptation of the additive heuristic for plan space planning, and modify it to account for possible reuse of existing actions in a plan. We also propose a large set of novel flaw selection strategies, and show how these can help us solve more problems than previously possible by POCL planners. VHPOP also supports planning with durative actions by incorporating standard techniques for temporal constraint reasoning. We demonstrate that the same heuristic techniques used to boost the performance of classical POCL planning can be effective in domains with durative actions as well. The result is a versatile heuristic POCL planner competitive with established CSP-based and heuristic state space planners.

## 1. Introduction

During the first half of the last decade, much of the research in domain independent plan generation focused on *partial order causal link* (POCL) planners. The two dominant POCL planners were SNLP (McAllester & Rosenblitt, 1991) and UCPOP (Penberthy & Weld, 1992), and a large part of the planning research was aimed at scaling up these two planners. The most promising attempts at making POCL planning practical involved alternative *flaw selection* strategies (Peot & Smith, 1993; Joslin & Pollack, 1994; Schubert & Gerevini, 1995; Williamson & Hanks, 1996; Pollack, Joslin, & Paolucci, 1997). A flaw in POCL planning is either an unlinked precondition (called *open condition*) for an action, or a threatened causal link. While flaw selection is not a backtracking point in the search through *plan space* for a complete plan, the order in which flaws are resolved can have a dramatic effect on the number of plans searched before a solution is found. The role of flaw selection in POCL planning is similar to the role of variable selection in constraint programming.

There have been dramatic advances in domain independent planning in the past seven years, but the focus has shifted from POCL planning to CSP-based planning algorithms (Blum & Furst, 1997; Kautz & Selman, 1996) and *state space* planning as heuristic search (Bonet & Geffner, 2001b; Hoffmann & Nebel, 2001). Recently, Nguyen and Kambhampati (2001) showed that with techniques such as distance based heuristics and reachability analysis—largely responsible for the efficiency of today's best domain independent planners—can also be used to dramatically improve the efficiency of POCL planners, thereby initiating a revival of this previously popular approach to domain independent planning. We have drawn from their experience, as well as from experience with flaw selection strategies from the glory-days of POCL planning, when developing the *Versatile Heuristic Partial Order Planner* (VHPOP), and the result is a POCL planner that was able to compete well with CSP-based and heuristic state space planners at the 3rd International Planning Competition (IPC3).

We have previously (Younes & Simmons, 2002) adapted the additive heuristic—proposed by Bonet, Loerincs, and Geffner (1997) and used in HSP (Bonet & Geffner, 2001b)—for plan space search. In this paper we present a variation of the additive heuristic for POCL planning that accounts for possible reuse of actions that are already part of a plan. We show that this accounting for positive interaction often results in a more effective plan ranking heuristic. We also present ablation studies that demonstrate the effectiveness of a tie-breaking heuristic based on estimated planning effort (defined as the total number of open conditions, current and future, that need to be resolved in order to complete a partial plan). The results show that using this tie-breaking heuristic almost always improves planner performance.

While the heuristics implemented in VHPOP can work with either ground (fully instantiated) or lifted (partially instantiated) actions, we chose to work only with ground actions at IPC3. We have shown elsewhere (Younes & Simmons, 2002) that planning with lifted actions can help reduce the branching factor of the search space compared to using ground actions, and that this reduction sometimes is large enough to compensate for the added complexity that comes with having to keep track of variable bindings. Further studies are needed, however, to gain a better understanding of the circumstances under which planning with lifted actions is beneficial.

VHPOP efficiently implements all the common flaw selection strategies, such as DUnf and DSep (Peot & Smith, 1993), LCFR (Joslin & Pollack, 1994), and ZLIFO (Schubert & Gerevini, 1995). In addition to these, we introduce numerous novel flaw selection strategies in this paper, of which four were used at IPC3. While we do not claim to have resolved the issue of global versus local flaw selection—manifested by the conflicting claims made by Gerevini and Schubert (1996) on the one hand, and Pollack et al. (1997) on the other about the most efficient way to reduce the number of searched plans in POCL planning—we show that by combining ideas from both ZLIFO and LCFR we can get very efficient flaw selection strategies. Other novel flaw selection strategies introduced in this paper are based on heuristic cost, an idea previously explored by Ghallab and Laruelle (1994). We also introduce "conflict-driven" flaw selection strategies that aim to expose possible inconsistencies early in the search, and we show that strategies based on this idea can be effective in domains previously thought to be particularly difficult for POCL planners.

Ideally, we would like to have one single flaw selection strategy that dominates all other strategies in terms of number of solved problems. We have yet to discover such a universal strategy, so instead we use a technique previously explored by Howe, Dahlman, Hansen, Scheetz, and von Mayrhauser (1999) for combining the strengths of different planning algorithms. The idea is to run several planners concurrently, and Howe et al. showed that by doing so more problems can be solved than by running any single planner. In VHPOP we use the same basic POCL planning algorithm in all instances, but we use different flaw selection strategies concurrently.

VHPOP extends the capabilities of classical POCL planners by also supporting planning with durative actions. This is accomplished by adding a *simple temporal network* (STN) (Dechter, Meiri, & Pearl, 1991) to the regular plan representation of a POCL planner. The STN records temporal constraints between actions in a plan, and supersedes the simple ordering constraints usually recorded by POCL planners. The use of STNs permits actions with interval constraints on the duration (a feature that was not utilized by any of the domains at IPC3 that VHPOP could handle). The approach we take to temporal POCL planning is essentially the same as the constraint-based interval approach described by Smith, Frank, and Jónsson (2000), and similar techniques for handling durative actions in a POCL framework can be traced back at least to Vere's DEVISER (Vere, 1983). Our contribution to temporal POCL planning is demonstrating that the same heuristic techniques shown to boost the performance of classical POCL planning can also be effective in domains with durative actions, validating the feasibility of the POCL paradigm for temporal planning on a larger set of benchmark problems than has been done before.

## 2. Basic POCL Planning Algorithm

We briefly review how POCL planners work, and introduce the terminology used throughout this paper. For a thorough introduction to POCL planning, we refer the reader to the tutorial on least commitment planning by Weld (1994).

A (partial) plan can be represented by a tuple $\langle \mathcal{A}, \mathcal{L}, \mathcal{O}, \mathcal{B} \rangle$, where $\mathcal{A}$ is a set of actions, $\mathcal{L}$ a set of causal links, $\mathcal{O}$ a set of ordering constraints defining a partial order on the set $\mathcal{A}$, and $\mathcal{B}$ a set of binding constraints on the action parameters ($\mathcal{B} = \emptyset$ if ground actions are used). Each action $a$ is an instance of some action schema $A$ in the planning domain, and a plan can contain multiple instances of the same action schema. A causal link, $a_i \xrightarrow{q} a_j$, represents a commitment by the planner that precondition $q$ of action $a_j$ is to be fulfilled by an effect of action $a_i$.

An *open condition*, $\xrightarrow{q} a_i$, is a precondition $q$ of action $a_i$ that has not yet been linked to an effect of another action. An *unsafe link* (or *threat*) is a causal link, $a_i \xrightarrow{q} a_j$, whose condition $q$ unifies with the negation of an effect of an action that could possibly be ordered between $a_i$ and $a_j$. The set of *flaws* of a plan $\pi$ is the union of open conditions and unsafe links: $\mathcal{F}(\pi) = \mathcal{OC}(\pi) \cup \mathcal{UL}(\pi)$.

A POCL planner searches for a solution to a planning problem in the space of partial plans by trying to resolve all flaws in a plan. Algorithm 1 shows a generic procedure for POCL planning that given a planning problem returns a plan solving the problem (or failure if the given problem lacks a solution). A planning problem is a set of initial conditions $\mathcal{I}$ and a set of goals $\mathcal{G}$, and is represented by an initial plan with two dummy actions $a_0 \prec a_\infty$, where the effects of $a_0$ represent the initial conditions of the problem and the preconditions of $a_\infty$ represent the goals of the problem. The procedure MAKE-INITIAL-PLAN used in Algorithm 1 returns the plan $\langle \{a_0, a_\infty\}, \emptyset, \{a_0 \prec a_\infty\}, \emptyset \rangle$. A set $\mathcal{P}$ of generated, but not yet visited, partial plans is kept. At each stage in the planning process, a plan is selected and removed from $\mathcal{P}$, and then a flaw is selected for that plan. All possible refinements resolving the flaw (returned by the procedure REFINEMENTS) are added to $\mathcal{P}$, and the process continues until $\mathcal{P}$ is empty (indicates failure) or a plan without flaws is found.

---

**Algorithm 1** Generic POCL planning algorithm as formulated by Williamson and Hanks (1996).

---

FIND-PLAN($\mathcal{I}, \mathcal{G}$)
  $\mathcal{P} \Leftarrow \{\text{MAKE-INITIAL-PLAN}(\mathcal{I}, \mathcal{G})\}$
  **while** $\mathcal{P} \neq \emptyset$ **do**
    $\pi \Leftarrow$ some element of $\mathcal{P}$   $\triangleright$ plan selection
    $\mathcal{P} \Leftarrow \mathcal{P} \setminus \{\pi\}$
    **if** $\mathcal{F}(\pi) = \emptyset$ **then**
      **return** $\pi$
    **else**
      $f \Leftarrow$ some element of $\mathcal{F}(\pi)$   $\triangleright$ flaw selection
      $\mathcal{P} \Leftarrow \mathcal{P} \cup \text{REFINEMENTS}(\pi, f)$
  **return** failure (problem lacks solution)

---

An open condition, $\xrightarrow{q} a_i$, can be resolved by linking $q$ to the effect of an existing or new action. An unsafe link, $a_i \xrightarrow{q} a_j$ threatened by the effect $p$ of action $a_k$, can be resolved by either ordering $a_k$ before $a_i$ (*demotion*), or by ordering $a_k$ after $a_j$ (*promotion*). If we use lifted actions instead of ground actions, a threat can also be resolved by adding binding constraints so that $p$ and $\neg q$ cannot be unified (*separation*).

## 3. Search Control

In the search for a complete plan, we first select a plan to work on, and given a plan we select a flaw to repair. These two choice points are indicated in Algorithm 1. Making an informed choice in both these cases is essential for good planner performance, and the following is a presentation of how these choices are made in VHPOP.

### 3.1 Plan Selection Heuristic

VHPOP uses the $A^*$ algorithm (Hart, Nilsson, & Raphael, 1968) to search through plan space. The $A^*$ algorithm requires a search node evaluation function $f(n) = g(n) + h(n)$, where $g(n)$ is the cost of getting to $n$ from the start node (initial plan) and $h(n)$ is the estimated remaining cost of reaching a goal node (complete plan). We want to find plans containing few actions, so we take the cost of a plan to be the number of actions in it. For a plan $\pi = \langle \mathcal{A}, \mathcal{L}, \mathcal{O}, \mathcal{B} \rangle$ we therefore have $g(\pi) = |\mathcal{A}|$.

The original implementations of SNLP and UCPOP used $h_{\mathrm{f}}(\pi) = |\mathcal{F}(\pi)|$ as the heuristic cost function, i.e. the number of flaws in a plan. Schubert and Gerevini (1995) consider alternatives for $h_{\mathrm{f}}(\pi)$, and present empirical data showing that just counting the open conditions ($h_{\mathrm{oc}}(\pi) = |\mathcal{OC}(\pi)|$) often gives better results. A big problem, however, with using the number of open conditions as an estimate of the number of actions that needs to be added is that it assumes a uniform cost per open condition. It ignores the fact that some open conditions can be linked to existing actions (thus requiring no additional actions), while other open conditions can be resolved only by adding a whole chain of actions (thus requiring more than one action).

Recent work in heuristic search planning has resulted in more informed heuristic cost functions for state space planners. We have in previous work (Younes & Simmons, 2002) adapted the additive heuristic—first proposed by Bonet et al. (1997) and subsequently used in HSP (Bonet & Geffner, 2001b)—for plan space search and also extended it to handle negated and disjunctive preconditions of actions as well as actions with conditional effects and lifted actions. The heuristic cost function used by VHPOP at IPC3 was a variation of the additive heuristic where some reuse of actions is taken into account, coupled with the tie-breaking rank (introduced in Younes & Simmons, 2002) based on estimated remaining planning effort.

### 3.1.1 The Additive Heuristic for POCL Planning

The key assumption behind the additive heuristic is subgoal independence. We give a recursive definition of the additive heuristic for POCL planning, starting at the level of literals and working towards a definition of heuristic cost for a partial plan.

Given a literal $q$, let $\mathcal{GA}(q)$ be the set of ground actions having an effect that unifies with $q$. The cost of the literal $q$ can then be defined as

$$h_{\text{add}}(q) = \begin{cases} 0 & \text{if } q \text{ unifies with a literal that holds initially} \\ \min_{a \in \mathcal{GA}(q)} h_{\text{add}}(a) & \text{if } \mathcal{GA}(q) \neq \emptyset \\ \infty & \text{otherwise} \end{cases} .$$

A positive literal $q$ holds initially if it is part of the initial conditions. A negative literal $\neg q$ holds initially if $q$ is not part of the initial conditions (the closed-world assumption). The cost of an action $a$ is

$$h_{\text{add}}(a) = 1 + h_{\text{add}}(Prec(a)),$$

where $Prec(a)$ is a propositional formula in *negation normal form* representing the preconditions of action $a$. A propositional formula is in negation normal form if negations only occur at the level of literals. Any propositional formula can be transformed into negation normal form, and this is done for action preconditions by VHPOP while parsing the domain description file.

Existentially quantified variables in an action precondition can be treated as additional parameters of the action. The cost of an existentially quantified precondition can then simply be defined as follows:

$$h_{\text{add}}(\exists x.\phi) = h_{\text{add}}(\phi)$$

We can deal with universally quantified preconditions by making them fully instantiated in a preprocessing phase, so in order to complete the definition of heuristic cost for action preconditions we only need to add definitions for the heuristic cost of conjunctions and disjunctions. The cost of a conjunction is the sum of the cost of the conjuncts:

$$h_{\text{add}}(\bigwedge_i \phi_i) = \sum_i h_{\text{add}}(\phi_i)$$

The summation in the above formula is what gives the additive heuristic its name. The definition is based on the assumption that subgoals are independent, which can lead to overestimation of the actual cost of a conjunctive goal (i.e. the heuristic is not admissible). The cost of a disjunction is taken to be the cost of the disjunct with minimal cost:

$$h_{\mathrm{add}}\Big(\bigvee_i \phi_i\Big) = \min_i h_{\mathrm{add}}(\phi_i)$$

The additive heuristic cost function for POCL plans can now be defined as follows:

$$h_{\mathrm{add}}(\pi) = \sum_{\xrightarrow{q} a_i \in \mathcal{OC}(\pi)} h_{\mathrm{add}}(q)$$

As with the cost function for conjunction, the above definition can easily lead to overestimation of the number of actions needed to complete a plan, since possible reuse is ignored. We propose a remedy for this below.

The cost of ground literals can be efficiently computed through dynamic programming. We take conditional effects into account in the cost computation. If the effect $q$ is conditioned by $p$ in action $a$, we add $h_{\mathrm{add}}(p)$ to the cost of achieving $q$ with $a$. We only need to compute the cost for ground literals once during a preprocessing phase, leaving little overhead for evaluating plans during the planning phase. When working with lifted actions, there is extra overhead for unification. It should also be noted that all lifted literals are independently matched to ground literals without considering interactions between open conditions of the same action. For example, two preconditions `(a ?x)` and `(b ?x)` of the same action can be unified to ground literals with different matchings for the variable `?x`.

### 3.1.2 Accounting for Positive Interaction

The additive heuristic does not take reuse of actions (other than the dummy action $a_0$) into account, so it often overestimates the actual number of actions needed to complete a plan. The need to take positive interaction into account in order to obtain a more accurate heuristic estimate has been recognized in both state space planning (Nguyen & Kambhampati, 2000; Hoffmann & Nebel, 2001; Refanidis & Vlahavas, 2001) and plan space planning (Nguyen & Kambhampati, 2001). For IPC3 we used a slight modification of the additive heuristic to address the issue of action reuse:

$$h^{\mathrm{r}}_{\mathrm{add}}(\pi) = \sum_{\xrightarrow{q} a_i \in \mathcal{OC}(\pi)} \begin{cases} 0 & \text{if } \exists a_j \in \mathcal{A} \text{ s.t. an effect of } a_j \text{ unifies with } q \\ & \quad \text{and } a_i \prec a_j \notin \mathcal{O} \\ h_{\mathrm{add}}(q) & \text{otherwise} \end{cases}$$

The underlying assumption for this heuristic cost function is that an open condition $\xrightarrow{q} a_i$ that can possibly be resolved by linking to the effect of an existing action $a_j$ will not give rise to a new action when resolved. This can of course lead to an overly optimistic estimate of the number of actions required to complete the plan. The modified heuristic is still not admissible, however, since the same cost value as before is used for open conditions that cannot be linked to effects of existing actions. In other words, we only account for possible reuse of existing actions and not potential actions.

To illustrate the difference between $h_{\mathrm{add}}(\pi)$ and $h^{\mathrm{r}}_{\mathrm{add}}(\pi)$ consider a planning domain with two action schemas $A_1$ and $A_2$, where $A_1$ has no preconditions and $A_2$ has a single precondition $q$. Assume that $q$ can only be achieved through an action instance of $A_1$. The heuristic cost for the literal $q$ is therefore 1 according to the additive heuristic. Consider now a plan $\pi$ with two unordered actions $a_1$ and $a_2$ ($a_i$ being an instance of action schema $A_i$) and a single open condition $\xrightarrow{q} a_2$. We have $h_{add}(\pi) = h_{add}(q) = 1$ corresponding to the addition of a new instance of action schema $A_1$ to achieve $q$, but $h_{add}^{r}(\pi) = 0$ because there is an action (viz. $a_1$) that is not ordered after $a_2$ and has an effect that unifies with $q$. Table 1 shows that taking reuse into account can have a significant impact on planning time in practice. The modified additive heuristic $h_{add}^{r}$ clearly dominates $h_{add}$ in the DriverLog and ZenoTravel domains despite incurring a higher overhead per generated plan. The results in the Satellite domain are more mixed, with $h_{add}$ having a slight edge overall. We show planning times for the four flaw selection strategies that were used by VHPOP at IPC3. These and other novel flaw selection strategies are discussed in detail in Section 3.2.2.

| Problem | MW-Loc | | MW-Loc-Conf | | LCFR-Loc | | LCFR-Loc-Conf | |
|---|---|---|---|---|---|---|---|---|
| | $h_{add}$ | $h_{add}^{r}$ | $h_{add}$ | $h_{add}^{r}$ | $h_{add}$ | $h_{add}^{r}$ | $h_{add}$ | $h_{add}^{r}$ |
| DriverLog6 | 8.65 | 0.16 | 4.41 | 0.13 | 87.58 | 2.01 | - | 1.16 |
| DriverLog7 | 3.66 | 0.34 | 0.63 | 0.17 | 21.15 | 1.28 | 1.57 | 0.22 |
| DriverLog8 | - | - | 110.26 | 1.48 | - | 177.27 | - | 2.05 |
| DriverLog9 | - | 0.33 | | 0.28 | | | - | - |
| DriverLog10 | 4.13 | 2.11 | 0.71 | 0.76 | 3.79 | 0.64 | 1.30 | 0.83 |
| ZenoTravel6 | - | 0.93 | 17.41 | 2.90 | 25.09 | 0.95 | 11.24 | 2.82 |
| ZenoTravel7 | - | - | - | 37.81 | - | - | - | 33.10 |
| ZenoTravel8 | - | 15.48 | - | 37.99 | - | - | - | 6.45 |
| ZenoTravel9 | - | 86.21 | - | 11.53 | - | 33.37 | 26.33 | 9.49 |
| ZenoTravel10 | - | 26.59 | - | 21.22 | - | 21.20 | - | 18.22 |
| Satellite6 | 0.36 | 0.22 | 0.37 | 0.24 | 0.32 | 0.21 | 0.40 | 0.24 |
| Satellite7 | 0.49 | 0.37 | 0.54 | 0.84 | 0.55 | 0.51 | 0.62 | - |
| Satellite8 | 1.09 | - | 1.29 | 0.84 | 0.85 | 0.83 | 1.25 | 0.68 |
| Satellite9 | 2.41 | - | 2.11 | - | 1.84 | - | 2.50 | - |
| Satellite10 | 1.53 | 1.12 | 1.95 | 1.11 | 1.50 | 1.36 | 2.08 | 1.37 |

Table 1: Planning times in seconds using different flaw selection strategies for a selection of problems in the DriverLog, ZenoTravel, and Satellite domains, showing the impact of taking reuse into account in the plan ranking heuristic. A dash (-) means that the planner ran out of memory (512 Mb).

Hoffmann and Nebel (2001) describe the FF heuristic that takes positive interaction between actions into account by extracting a plan from the *relaxed* planning graph[1], and argue that the accounting of action reuse is one of the contributing factors to FF's performance advantage over HSP. The FF heuristic can take reuse of potential actions into account, and not just existing actions as is the case with our modified additive heuristic. This should result in a better estimate of actual plan cost, but requires that a plan is extracted from the relaxed planning graph for every search node, which could be costly. It would be interesting to see how the FF heuristic performs if used in a plan space planner.

---

1. A relaxed planning graph is a planning graph with no action pairs marked as mutex.

The heuristic cost function used in REPOP (Nguyen & Kambhampati, 2001), a heuristic partial order planner working solely with ground actions, is defined using a *serial* planning graph.[2] The heuristic is similar in spirit to the FF heuristic, and can like the FF heuristic take reuse of potential actions into account. The REPOP heuristic also takes into account reuse of existing actions, but seemingly without considering ordering constraints, which is something we do in our modified additive heuristic. Furthermore, our $h_{\text{add}}^{\text{r}}$ heuristic always takes reuse of any existing actions that achieves a literal $q$ into account, while the REPOP heuristic only considers an existing action if it happens to be selected from the serial planning graph as the the action that achieves $q$. The results in Table 2 indicate that the REPOP heuristic may be less effective than the additive heuristic (with and without reuse) in certain domains.

### 3.1.3 Estimating Remaining Effort

Not only do we want to find plans consisting of few actions, but we also want to do so exploring as few plans as possible. Schubert and Gerevini (1995) suggest that the number of open conditions can be useful as an estimate of the number of refinement steps needed to complete a plan. We take this idea a bit further.

When computing the heuristic cost of a literal, we also record the estimated effort of achieving the literal. A literal that is achieved through the initial conditions has estimated effort 1 (corresponding to the work of adding a causal link to the plan). If the cost of a literal comes from an action $a$, the estimated effort for the literal is the estimated effort for the preconditions of $a$, plus 1 for linking to $a$. Finally, the estimated effort of a conjunction is the sum of the estimated effort of the conjuncts, while the estimated effort of a disjunction is the estimated effort of the disjunct with minimal *cost* (not effort).

The main difference between heuristic cost and estimated effort of a plan is that estimated effort assigns the value 1 instead of 0 to literals that can be unified with an initial condition. To illustrate the difference, consider a plan $\pi$ with two open conditions $p$ and $q$ that both hold in the initial conditions. The heuristic cost for $\pi$ is 0, while the estimated effort is 2. The estimated effort is basically a heuristic estimate of the total number of open conditions that will have to be resolved before a complete plan is found, and it is used as a tie-breaker between two plans $\pi$ and $\pi'$ in case $f(\pi) = f(\pi')$. Consider an alternative plan $\pi'$ with the same number of actions as $\pi$ but with a single open condition $p$. This plan has heuristic cost 0 as does the plan $\pi$, but the estimated effort is only 1, so $\pi'$ would be selected first if estimated effort is used as a tie-breaker. Table 2 shows that using estimated effort as a tie-breaker can have a notable impact on planner performance for both $h_{\text{add}}$ and $h_{\text{add}}^{\text{r}}$. Estimated effort helps reduce the number of generated and explored plans in all cases but one (when using $h_{\text{add}}^{\text{r}}$ on problem rocket-ext-a).

Estimated effort is not only useful as a plan ranking heuristic, but also for heuristic flaw selection as we will soon see.

---

2. A serial planning graph is a planning graph with every pair of non-noop actions at the same level marked as mutex.

| Problem | $h_{\text{add}}$ | with effort | $h_{\text{add}}^{\text{r}}$ | with effort | REPOP |
|---|---|---|---|---|---|
| gripper-8 | 1636 / 705 | 1089 / 449 | * | * | * |
| gripper-10 | 3268 / 1359 | 1958 / 795 | * | * | * |
| gripper-12 | 5879 / 2359 | 3224 / 1294 | * | * | * |
| gripper-20 | 33848 / 12204 | 14386 / 5558 | * | * | * |
| rocket-ext-a | 34917 / 25810 | 27846 / 20028 | 24507 / 15790 | 31213 / 20321 | 30110 / 17768 |
| rocket-ext-b | 27871 / 20034 | 27277 / 19363 | 15919 / 9000 | 10914 / 6705 | 85316 / 51540 |
| logistics-a | 503 / 301 | 481 / 287 | 621 / 389 | 530 / 317 | 411 / 191 |
| logistics-b | 857 / 488 | 713 / 404 | 694 / 402 | 584 / 326 | 920 / 436 |
| logistics-c | 766 / 422 | 630 / 346 | 629 / 353 | 438 / 227 | 4939 / 2468 |
| logistics-d | 3039 / 1398 | 2950 / 1384 | 2525 / 1300 | 1472 / 682 | * |

Table 2: The number of generated/explored plans for $h_{\text{add}}$ and $h_{\text{add}}^{\text{r}}$ both without and with estimated effort as a tie-breaker. The REPOP column contains the numbers reported by Nguyen and Kambhampati (2001) for REPOP using the serial planning graph heuristic. These numbers are included only for the purpose of showing that there seems to be a qualitative difference between the REPOP heuristic and the heuristics used by VHPOP. An asterisk (*) means that no solution was found after generating 100000 plans. Flaws were selected in LIFO order.

## 3.2 Flaw Selection Strategies

In the original implementations of SNLP and UCPOP threats are selected before open conditions. When there is more than one threat (or open condition) that can be selected, the one added last is selected first (LIFO order). Several alternative flaw selection strategies have been proposed in an attempt to improve the performance exhibited by POCL planners.

Peot and Smith (1993) show that the number of searched plans can be reduced by delaying the resolution of some threats. The most successful of the proposed delay strategies are DSep, which delays threats that can be resolved through separation, and DUnf, which delays threats that can be resolved in more than one way.

Joslin and Pollack (1994) suggest that all flaws should be treated uniformly, and that the flaw with the least number of refinements should be selected first. Their flaw selection strategy, LCFR, can be viewed as an instance of the most-constrained-variable heuristic used in simple search rearrangement backtracking (Bitner & Reingold, 1975; Purdom, 1983). The main disadvantage with LCFR is that computing the repair cost for every flaw can incur a large overhead for flaw selection. This can lead to longer planning times compared to when using the default UCPOP strategy, even if the number of search nodes is significantly smaller with LCFR. A clever implementation of LCFR can, however, reduce the overhead for flaw selection considerably.

Schubert and Gerevini (1995) argue that a LIFO strategy for selecting open conditions helps the planner maintain focus on the achievement of a particular high-level goal. Their ZLIFO strategy is a variation of the DSep strategy, with the difference being that open conditions that cannot be resolved, or can be resolved in only one way, are selected before open conditions that can be resolved in more than one way. Gerevini and Schubert (1996) present results indicating that ZLIFO often needs to generate fewer plans than LCFR before a solution is found, and has a smaller overhead for flaw selection. These results are disputed by Pollack et al. (1997). They instead attribute much of the power of ZLIFO to its delaying of separable threats, and propose a variation of LCFR, LCFR-DSep, that also delays separable threats. Since we chose to work with ground actions at IPC3, separability was not an issue for us.

### 3.2.1 NOTATION FOR SPECIFYING FLAW SELECTION STRATEGIES

In order to better understand the differences between various flaw selection strategies, and to simplify comparative studies, Pollack et al. (1997) proposed a unifying notation for specifying flaw selection strategies. We adopt their notation with only slight modifications.

A flaw selection strategy is an ordered list of selection criteria. Each selection criterion is of the form

$$\{\textit{flaw types}\}_{\leq \textit{max refinements}} \textit{ordering criterion},$$

and applies to flaws of the given types that can be resolved in at most *max refinements* ways. If there is no limit on the number of refinements, we simply write

$$\{\textit{flaw types}\} \textit{ordering criterion}.$$

The ordering criterion is used to order flaws that the selection criterion applies to. LIFO order is used if the ordering criterion cannot be used to distinguish two or more flaws.

Pollack et al. define the flaw types "o" (open condition), "n" (non-separable threat), and "s" (separable threat). They also define the ordering criteria "LIFO", "FIFO", "R" (random), "LR"[3] (least refinements first), and "New". The last one applies only to open conditions, and gives preference to open conditions that can be resolved by adding a new action. The rest apply to both open conditions and threats.

Flaws are matched with selection criteria, and it is required for completeness that every flaw matches at least one selection criterion in a flaw selection strategy. The flaw that matches the earliest selection criterion, and is ordered before any other flaws matching the same criterion (according to the ordering criterion), is the flaw that gets selected by the flaw selection strategy. Note that we do not always need to test all flaws. If, for example, the first selection criterion is {n, s}LIFO, and we have found a threat, then we do not need to consider any other flaws for selection.

Using this notation, we can specify many different flaw selection strategies in a concise manner. Table 3 specifies the flaw selection strategies mentioned earlier. A summary of flaw types recognized by VHPOP, including three new flaw types defined below, is given in Table 4.

### 3.2.2 NEW FLAW SELECTION STRATEGIES

We now propose several additional flaw types and ordering criteria, and use these in combination with the previous ones to obtain some novel flaw selection strategies. Four of these new flaw selection strategies were used at IPC3 and contributed to the success of VHPOP at that event.

---

3. The original notation for this ordering criterion is LC for "least (repair) cost", where the repair cost is defined to be the number of refinements. Because we introduce a new ordering criterion based on heuristic cost, we choose to rename this ordering criterion.

| Name | Specification |
|------|---------------|
| UCPOP | {n, s}LIFO / {o}LIFO |
| DSep | {n}LIFO / {o}LIFO / {s}LIFO |
| DUnf | $\{n,s\}_{\leq 0}$LIFO / $\{n,s\}_{\leq 1}$LIFO / {o}LIFO / {n, s}LIFO |
| LCFR | {n, s, o}LR |
| LCFR-DSep | {n, o}LR / {s}LR |
| ZLIFO | {n}LIFO / $\{o\}_{\leq 0}$LIFO / $\{o\}_{\leq 1}$New / {o}LIFO / {s}LIFO |

Table 3: A few of the flaw selection strategies previously proposed in the planning literature.

| Flaw Type | Description |
|-----------|-------------|
| n | non-separable threat |
| s | separable threat |
| o | open condition |
| t | static open condition |
| l | local open condition |
| u | unsafe open condition |

Table 4: Summary of flaw types recognized by VHPOP.

**Early Commitment through Flaw Selection.** We have shown (Younes & Simmons, 2002) that giving priority to *static open conditions* can be beneficial when planning with lifted actions. Introducing a new flaw type, "t", representing static open conditions, we can specify this flaw selection strategy as follows:

**Static-First** {t}LIFO / {n, s}LIFO / {o}LIFO

A static open condition is a literal that involves a predicate occurring in the initial conditions of a planning problem, but not in the effects of any operator in the planning domain. This means that a static open condition always has to be linked to the initial conditions, and the initial conditions consist solely of ground literals. Resolving a static open condition $\xrightarrow{q} a_i$ will therefore cause all free variables of $q$ to be bound to specific objects. Resolving static open conditions before other flaws represents a bias towards early commitment of parameter bindings. This resembles the search strategy inherent in planners using ground actions, but without necessarily committing to bindings for all parameters of an action at once. The gain is a reduced branching factor compared to a planner using ground actions, and this reduction can compensate for the increased complexity that comes with having to keep track of variable bindings.

Our earlier results (Younes & Simmons, 2002) indicated that despite a reduction in the number of generated plans when planning with lifted actions, using ground actions was still faster in most domains. In the gripper domain, for example, while using lifted actions resulted in less than half the number of generated plans compared to when using ground actions, planning with ground actions was still more than twice as fast. We have greatly improved the implementation of the planner and the handling of variable bindings since then. When using the latest version of VHPOP in the gripper domain, the planner is roughly as fast when planning with lifted actions giving priority to static preconditions as when planning with ground actions.

**Local Flaw Selection.** By retaining the LIFO order for selecting open conditions achievable in multiple ways, Schubert and Gerevini (1995) argue that the planner tends to maintain focus on a particular higher-level goal by regression, instead of trying to achieve multiple goals in a breadth-first manner. When some of the goals to achieve are independent, maintaining focus on a single goal should be beneficial. The problem with a LIFO-based flaw selection strategy, however, as pointed out by Williamson and Hanks (1996), is that it is highly sensitive to the order in which operator preconditions are specified in the domain description.

It is not necessary, however, to select the most recently added open condition in order to keep focus on the achievement of one goal. We can get the same effect by selecting *any* of the open conditions, but restrict the choice to the most recently added action. We therefore introduce a new flaw type, "l", representing *local open conditions*. A local open condition is one that belongs to the most recently added action that still has remaining open conditions. We can use any ordering criterion to select among local open conditions. Using this new flaw type, we can specify a local variant of LCFR:

$$\textbf{LCFR-Loc} \quad \{n, s, l\}LR$$

One would expect such a strategy to be less sensitive to precondition-order than a simple LIFO-based strategy. We can see evidence of this in Table 5, which also shows that the maintained goal focus achieved by local flaw selection strategies can help solve more problems compared to a global flaw selection strategy.

**Heuristic Flaw Selection.** Distance based heuristics have been used extensively for ranking plans in state space planners (e.g., HSP and FF). Nguyen and Kambhampati (2001) show that these heuristics can be very useful for ranking plans in POCL planners as well. They also suggest that the same heuristics could be used in flaw selection methods, but do not elaborate further on this subject.

It is not hard to see, however, how many of the plan rank heuristics could be used for the purpose of selecting among open conditions, since they often are based on estimating the cost of achieving open conditions as seen in Section 3.1.1. By giving priority to open conditions with the highest heuristic cost, we can build plans in a top-down manner from the goals to the initial conditions. We call this ordering criterion "MC" (most cost first). By using the opposite ordering criterion, "LC", we would instead tend to build plans in a bottom-up manner. Note that these two ordering criteria only apply to open conditions, and not to threats, so we would need to use them in combination with selection criteria for threats. We can define both global and local heuristic flaw selection strategies:

$$\textbf{MC} \quad \{n, s\}LR \; / \; \{o\}MC_{add}$$
$$\textbf{MC-Loc} \quad \{n, s\}LR \; / \; \{l\}MC_{add}$$

The subscript for MC indicates the heuristic function to use for ranking open conditions, which in this case is the additive heuristic.

| Problem | UCPOP | | LCFR | | LCFR-Loc | | MC | | MC-Loc | | MW | | MW-Loc | |
|---|---|---|---|---|---|---|---|---|---|---|---|---|---|---|
| | $\sigma/|\mu|$ | $n$ | $\sigma/|\mu|$ | $n$ | $\sigma/|\mu|$ | $n$ | $\sigma/|\mu|$ | $n$ | $\sigma/|\mu|$ | $n$ | $\sigma/|\mu|$ | $n$ | $\sigma/|\mu|$ | $n$ |
| DriverLog6 | 0.20 | 20 | 0.01 | 20 | 0.01 | 20 | 0.18 | 20 | 0.23 | 20 | 0.02 | 20 | 0.02 | 20 |
| DriverLog7 | 0.23 | 20 | 0.10 | 20 | 0.32 | 20 | 0.13 | 18 | 0.25 | 20 | - | 0 | 0.05 | 20 |
| DriverLog8 | 0.28 | 17 | - | 0 | 0.00 | 1 | - | 0 | - | 0 | - | 0 | - | 0 |
| DriverLog9 | 0.62 | 7 | 0.00 | 10 | 0.45 | 14 | - | 0 | 0.01 | 20 | - | 0 | 0.01 | 20 |
| DriverLog10 | 0.33 | 16 | - | 0 | 0.07 | 20 | - | 0 | 0.08 | 20 | - | 0 | 0.08 | 20 |
| ZenoTravel6 | 0.27 | 20 | 0.03 | 7 | 0.22 | 20 | - | 0 | 0.00 | 20 | - | 0 | 0.00 | 20 |
| ZenoTravel7 | 0.23 | 8 | - | 0 | 0.18 | 16 | - | 0 | 0.16 | 16 | - | 0 | 0.16 | 16 |
| ZenoTravel8 | 0.29 | 11 | - | 0 | 0.15 | 19 | - | 0 | 0.18 | 20 | - | 0 | 0.18 | 20 |
| ZenoTravel9 | 0.22 | 17 | - | 0 | 0.21 | 18 | - | 0 | 0.19 | 20 | - | 0 | 0.19 | 20 |
| ZenoTravel10 | 0.26 | 18 | - | 0 | 0.22 | 17 | - | 0 | 0.15 | 19 | - | 0 | 0.15 | 19 |
| Satellite6 | 0.20 | 19 | - | 0 | 0.02 | 20 | - | 0 | 0.02 | 20 | - | 0 | 0.02 | 20 |
| Satellite7 | 0.54 | 9 | - | 0 | 0.03 | 20 | - | 0 | 0.07 | 20 | - | 0 | 0.07 | 20 |
| Satellite8 | 0.35 | 8 | - | 0 | 0.02 | 20 | - | 0 | 0.07 | 4 | - | 0 | 0.07 | 4 |
| Satellite9 | 0.34 | 7 | - | 0 | 0.00 | 1 | - | 0 | 0.00 | 1 | - | 0 | 0.00 | 1 |
| Satellite10 | 0.32 | 9 | - | 0 | 0.01 | 20 | - | 0 | - | 0 | - | 0 | - | 0 |

Table 5: Relative standard deviation for the number of generated plans $(\sigma/|\mu|)$ and the number of solved problems $(n)$ over 20 instances of each problem with random precondition ordered. Low relative standard deviation indicates low sensitivity to precondition order. Results are shown for VHPOP using seven different flaw selection strategies. A memory limit of 512 Mb was enforced, and $h_{add}^r$ with estimated effort as tie-breaker was used as plan ranking heuristic.

In Section 3.1.3, we proposed that we can estimate the remaining planning effort for an open condition by counting the total number of open conditions that would arise while resolving the open condition. This heuristic could also be useful for ranking open conditions, and is often more discriminating than an ordering criterion based on heuristic cost. We therefore define two additional ordering criteria: "MW" (most work first) and "LW". With these, we can define additional flaw selection strategies:

$$\textbf{MW} \quad \{n, s\}LR \ / \ \{o\}MW_{add}$$
$$\textbf{MW-Loc} \quad \{n, s\}LR \ / \ \{l\}MW_{add}$$

For the planning problems listed in Table 5, we can see that MW-Loc is at most as sensitive to precondition order as MC-Loc, with MW-Loc never performing worse than MC-Loc and for the first two problems performing clearly better.

IxTeT (Ghallab & Laruelle, 1994; Laborie & Ghallab, 1995) also uses heuristic techniques to guide flaw selection, but in quite a different way than suggested here. It is our understanding of the IxTeT heuristic that it estimates, for each possible refinement $r$ resolving a flaw, the amount of change (commitment) that would result from applying $r$ to the current plan. For open conditions, this estimate is obtained by expanding a tree of subgoal decomposition, which in principal is a regression-match graph (McDermott, 1999). This is similar to how heuristic values are computed using the additive heuristic. However, IxTeT considers not only the number of actions that need to be added to resolve an open condition but also to what degree current variable domains would be reduced and possible action orderings restricted. Furthermore, IxTeT uses the heuristic values to choose the flaw in which a single refinement stands out the most as the least "costly" compared to other refinements for the same flaw. The intended effect is a reduction in the amount of backtracking that is needed to find a solution, although we are not aware of any evaluation of the effectiveness of the technique.

**Conflict-Driven Flaw Selection.** Common wisdom in implementing search heuristics for constraint satisfaction problems, e.g. propositional satisfiability, is to first make decisions with maximal consequences, so that inconsistencies can be detected early on, pruning large parts of the search space.

A flaw selection strategy that follows this principle would be to link *unsafe open conditions* before other open conditions. We call an open condition unsafe if a causal link to that open condition would be threatened. By giving priority to unsafe open conditions, the planner will direct attention to possible conflicts/inconsistencies in the plan at an early stage. We introduce the flaw type "u" representing unsafe open conditions. Examples of conflict-driven flaw selection strategies using this new flaw type are the following variations of LCFR, LCFR-Loc, and MW-Loc:

$$\textbf{LCFR-Conf} \quad \{n, s, u\}LR \; / \; \{o\}LR$$
$$\textbf{LCFR-Loc-Conf} \quad \{n, s, u\}LR \; / \; \{l\}LR$$
$$\textbf{MW-Loc-Conf} \quad \{n, s\}LR \; / \; \{u\}MW_{add} \; / \; \{l\}MW_{add}$$

The first two of these conflict-driven strategies are very effective in the *link-chain* domain constructed by Veloso and Blythe (1994). The link-chain domain is an artificial domain specifically constructed to demonstrate the weakness of POCL planners in certain domains. What makes the domain hard for SNLP and UCPOP with their default flaw selection strategies is that open conditions can be achieved by several actions but with only one action being the right choice because of negative interaction. This forces the POCL planner to backtrack excessively over link commitments, but inconsistencies may not be immediately detected because of the many link alternatives. We can see in Figure 1 that VHPOP using the UCPOP flaw selection strategy performs very poorly in the link-chain domain. Using a more sophisticated flaw selection strategy such as LCFR improves performance somewhat. However, with the two conflict-driven flaw selection strategies all problems are solved in less than a second. The number of generated and explored plans is in fact identical for LCFR-Conf and LCFR-Loc-Conf, but LCFR-Loc-Conf is roughly twice as fast as LCFR-Conf because of reduced overhead. This demonstrates the benefit of local flaw selection strategies. Note, however, that LCFR is faster than LCFR-Loc in the link-chain domain, so local strategies are not always superior to global strategies.

We can also see in Table 1 that conflict-driven flaw selection strategies work well in the DriverLog and Depots domains, both with $h_{add}$ and $h_{add}^r$ as heuristic function for ranking plans.

## 4. Temporal POCL Planning

In classical planning, actions have no duration: the effects of an action are instantaneous. Many realistic planning domains, however, require actions that can overlap in time and have different duration. The version of the planning domain definition language (PDDL), PDDL2.1, that was used for IPC3 introduces the notion of *durative actions*. A durative action represents an interval of time, and conditions and effects can be associated with either endpoint of this interval. Durative actions can also have invariant conditions that must hold for the entire duration of the action.

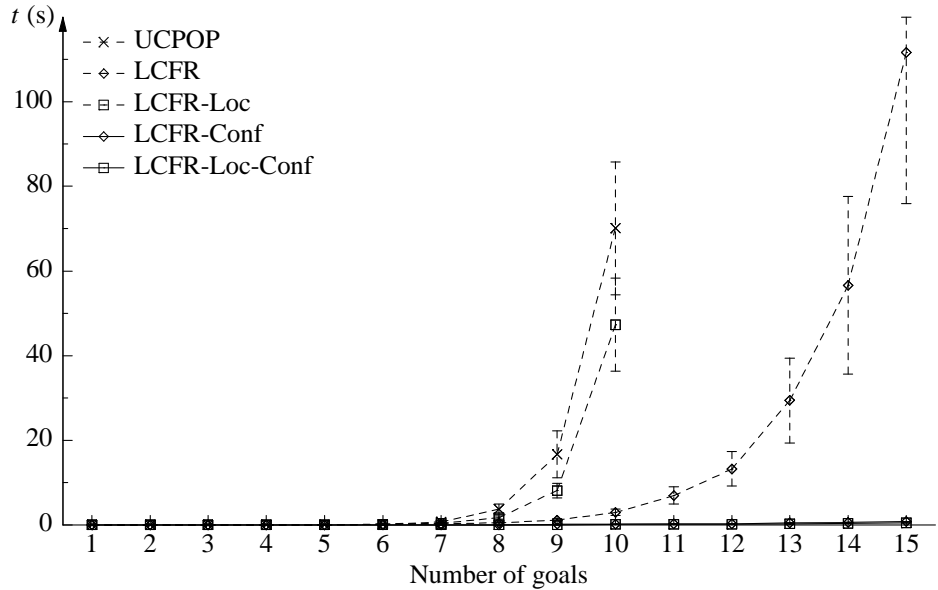

Figure 1: Average planning time over ten problems for each point with different flaw selection strategies in the link-chain domain. Results are shown for VHPOP using five different flaw selection strategies. Only points for which a strategy solved all ten problems without running out of memory (512 Mb) are shown. The $h_{\mathrm{f}}$ heuristic was used to rank plans.

We use the *constraint-based interval* approach to temporal POCL planning described by Smith et al. (2000), which in essence is the same approach as used by earlier temporal POCL planners such as DEVISER (Vere, 1983), ZENO (Penberthy & Weld, 1994), and IxTeT (Ghallab & Laruelle, 1994). Like IxTeT, we use a simple temporal network (STN) to record temporal constraints. The STN representation allows for rapid response to temporal queries. ZENO, on the other hand, uses an integrated approach for handling both temporal and metric constraints, and does not make use of techniques optimized for temporal reasoning. The following is a description of how VHPOP handles the type of temporal planning domains expressible in PDDL2.1.

When planning with durative actions, we substitute the partial order $\mathcal{O}$ in the representation of a plan with an STN $\mathcal{T}$. Each action $a_i$ of a plan, except the dummy actions $a_0$ and $a_\infty$, is represented by two nodes $t_{2i-1}$ (start time) and $t_{2i}$ (end time) in the STN $\mathcal{T}$, and $\mathcal{T}$ can be compactly represented by the *d-graph* (Dechter et al., 1991). The d-graph is a complete directed graph, where each edge $t_i \to t_j$ is labeled by the shortest temporal distance, $d_{ij}$, between the two time nodes $t_i$ and $t_j$ (i.e. $t_j - t_i \leq d_{ij}$). An additional time

$$\begin{pmatrix} 0 & \infty & \infty \\ \mathbf{-1} & 0 & \mathbf{7} \\ -4 & \mathbf{-3} & 0 \end{pmatrix}$$

(a)

$$\begin{pmatrix} 0 & \infty & \infty & \infty & \infty \\ -1 & 0 & 7 & \infty & \infty \\ -4 & -3 & 0 & \infty & \infty \\ \mathbf{-1} & \infty & \infty & 0 & \mathbf{4} \\ -5 & \infty & \infty & \mathbf{-4} & 0 \end{pmatrix}$$

(b)

$$\begin{pmatrix} 0 & \infty & \infty & \infty & \infty \\ -1 & 0 & 7 & 2 & 6 \\ -6 & -3 & 0 & -5 & \mathbf{-1} \\ -1 & \infty & \infty & 0 & 4 \\ -5 & \infty & \infty & -4 & 0 \end{pmatrix}$$

(c)

Figure 2: Matrix representation of d-graph, with $\epsilon = 1$, for STN after (a) adding action $a_1$ with duration constraint $\delta_1 \leq 7 \wedge \delta_1 \geq 3$, (b) adding action $a_2$ with duration constraint $\delta_2 = 4$, and (c) ordering the end of $a_2$ before the end of $a_1$. Explicitly added temporal constraints are in boldface.

point, $t_0$, is used as a reference point to represent time zero. By default, $d_{ij} = \infty$ for all $i \neq j$ ($d_{ii} = 0$), signifying that there is no upper bound on the difference $t_j - t_i$.

Constraints are added to $\mathcal{T}$ at the addition of a new action, the linking of an open condition, and the addition of an ordering constraint between endpoints of two actions.

The duration, $\delta_i$, of a durative action $a_i$ is specified as a conjunction of simple duration constraints $\delta_i \bowtie c$, where $c$ is a real-valued constant and $\bowtie$ is in the set $\{=, \leq, \geq\}$.[4] Each simple duration constraint gives rise to temporal constraints between the time nodes $t_{2i-1}$ and $t_{2i}$ of $\mathcal{T}$ when adding $a_i$ to a plan $\langle \mathcal{A}, \mathcal{L}, \mathcal{T}, \mathcal{B} \rangle$. The temporal constraints, in terms of the minimum distance $d_{ij}$ between two time points, are as follows:

| Duration Constraint | Temporal Constraints |
|---|---|
| $\delta_i = c$ | $d_{2i-1,2i} = c$ and $d_{2i,2i-1} = -c$ |
| $\delta_i \leq c$ | $d_{2i-1,2i} \leq c$ |
| $\delta_i \geq c$ | $d_{2i,2i-1} \leq -c$ |

The semantics of PDDL2.1 with durative actions dictates that every action be scheduled strictly after time zero. Let $\epsilon$ denote the smallest fraction of time required to separate two time points. To ensure that an added action $a_i$ is scheduled after time zero, we add the temporal constraint $d_{2i-1,0} \leq -\epsilon$ in addition to any temporal constraints due to duration constraints. Figure 2(a) shows the matrix representation of the d-graph after adding an action, $a_1$, with duration constraint $\delta_1 \leq 7 \wedge \delta_1 \geq 3$ to a null plan. The rows and columns of the matrix correspond to time point 0, the start of action $a_1$, and the end of action $a_1$ in that order. After adding action $a_2$ with duration constraint $\delta_2 = 4$, we have the d-graph represented by the matrix in Figure 2(b). The two additional rows and columns correspond to the start and end of action $a_2$ in that order.

A temporal annotation $\tau \in \{\text{s}, \text{i}, \text{e}\}$ is added to the representation of open conditions. The open condition $\xrightarrow{q@\text{s}} a_i$ represents a condition that must hold at the start of the durative action $a_i$, $\xrightarrow{q@\text{e}} a_i$ represents a condition that must hold at the end of $a_i$, while $\xrightarrow{q@\text{i}} a_i$ is an invariant condition for $a_i$. An equivalent annotation is added to the representation of causal links. The linking of an open condition $\xrightarrow{q@\tau} a_i$ to an effect associated with a time point $t_j$ gives rise to the temporal constraint $d_{kj} \leq -\epsilon$ ($k = 2i$ if $\tau = \text{e}$, else $k = 2i - 1$). Figure 2(c) shows the representation of the STN for a plan with actions $a_1$ and $a_2$, as before, and with an effect associated with the end of $a_2$ linked to a condition associated with the end of $a_1$.

---

4. In contrast, Vere's DEVISER can only handle duration constraints of the form $\delta_i = c$.

Unsafe causal links are resolved in basically the same way as before, but instead of adding ordering constraints between actions we add temporal constraints between time points ensuring that one time point precedes another time point. We can ensure that time point $t_i$ precedes time point $t_j$ by adding the temporal constraint $d_{ji} \leq -\epsilon$.

Every time we add a temporal constraint to a plan, we update all shortest paths $d_{ij}$ that could have been affected by the added constraint. This propagation of constraints can be carried out in $O(|\mathcal{A}|^2)$ time.

Once a plan without flaws is found, we need to schedule the actions in the plan, i.e. assign a start time and duration for each action. A schedule of the actions is a solution to the STN $\mathcal{T}$, and a solution assigning the earliest possible start time to each action is readily available in the d-graph representation. The start time of action $a_i$ is set to $-d_{2i-1,0}$ (Corollary 3.2, Dechter et al., 1991) and the duration to $d_{2i-1,0} - d_{2i,0}$. Assuming Figure 2(c) represents the STN for a complete plan, then we would schedule $a_1$ at time 1 with duration 5 and $a_2$ at time 1 with duration 4. We can easily verify that this schedule is indeed consistent with the duration constraints given for the actions, and that $a_2$ ends before $a_1$ as required.

Each non-durative action can be treated as a durative action of fixed duration 0, with preconditions associated with the start time, effects associated with the end time, and without any invariant conditions. This allows for a frictionless treatment of domains with both durative and non-durative actions.

Let us for a moment consider the memory requirements for temporal POCL planning compared to classical POCL planning. When planning with non-durative actions, we store $\mathcal{O}$ as a bit-matrix representing the transitive closure of the ordering constraints in $\mathcal{O}$. For a partial plan with $n$ actions, this requires $n^2$ bits. With $n$ durative actions, on the other hand, we need roughly $4n^2$ floating-point numbers to represent the d-graph of $\mathcal{T}$. Each floating-point number requires at least 32 bits on a modern machine, so in total we need more than 100 times as many bits to represent temporal constraints as regular ordering constraints for each plan. We note, however, that each refinement changes only a few entries in the d-graph, and by choosing a clever representation of matrices we can share storage between plans. The upper left $3 \times 3$ sub-matrix in Figure 2(b) is for example identical to the matrix in Figure 2(a). The way we store matrices in VHPOP allows us to exploit this commonality and thereby reduce the total memory requirements.

The addition of durative actions does not change the basic POCL algorithm. The recording of temporal constraints and temporal annotations can be handled in a manner transparent to the rest of the planner. The search heuristics described in Section 3, although not tuned specifically for temporal planning, can be used with durative actions. We only need to slightly modify the definition of literal and action cost in the additive heuristic because of the temporal annotations associated with preconditions and effects of durative actions. Let $\mathcal{GA}_\text{s}(q)$ denote the set of ground actions achieving $q$ at the start, and $\mathcal{GA}_\text{e}(q)$

| Name | Specification |
|------|---------------|
| MW-Loc | $\{n, s\}$LR / $\{l\}$MW$_{add}$ |
| MW-Loc-Conf | $\{n, s\}$LR / $\{u\}$MW$_{add}$ / $\{l\}$MW$_{add}$ |
| LCFR-Loc | $\{n, s, l\}$LR |
| LCFR-Loc-Conf | $\{n, s, u\}$LR / $\{l\}$LR |

Table 6: Flaw selection strategies used by VHPOP at IPC3.

the set of ground actions achieving $q$ at the end. We define the cost of the literal $q$ as

$$h_{add}(q) = \begin{cases} 0 & \text{if } q \text{ holds initially} \\ \min_{a \in \mathcal{GA}_t(q)} h_{add}(a@t) & \text{if } \mathcal{GA}_t(q) \neq \emptyset \\ \infty & \text{otherwise} \end{cases},$$

with $t \in \{s, e\}$ and the cost of a durative action $a$ at endpoint $t$ defined as

$$h_{add}(a@t) = 1 + h_{add}(Prec_t(a)).$$

$Prec_s(a)$ is a propositional formula representing the invariant preconditions of $a$ and preconditions associated with the start of $a$, while $Prec_e(a)$ is a formula representing *all* preconditions of $a$.

## 5. VHPOP at IPC3

VHPOP allows for several flaw selection strategies to be used simultaneously in a round-robin scheme. This lets us exploit the strengths of different flaw selection strategies concurrently, which was essential for the success of VHPOP at IPC3 since we have yet to find a single superior flaw selection strategy that dominates all other flaw selection strategies in terms of the number of solved problems within a given time frame. The technique we use in VHPOP for supporting multiple flaw selection strategies is in essence the same as the technique proposed by Howe et al. (1999) for exploiting performance benefits of several planners at once in a meta-planner. Although the meta-planner is slower than the fastest planner on any single problem, it can solve more problems than any single planner.

We used four different flaw selection strategies at IPC3 (Table 6), preferring local flaw selection strategies as they tend to incur a lower overhead than global strategies such as LCFR and MW and often appear more effective than global strategies because of a maintained focused on subgoal achievement. The four strategies were selected after some initial experimentation with problems from a few of the competition domains.

The use of multiple flaw selection strategies can be thought of as running multiple concurrent instances of the planner, as a separate search queue is maintained for each flaw selection strategy that is used. Similar to HSP2.0 at the planning competition in 2000 (Bonet & Geffner, 2001a), we use a fixed control strategy to schedule these multiple instances of our planner. The first time a flaw selection strategy is used, it is allowed to generate up to 1000 search nodes. The second time the same flaw selection strategy is used, it can generate another 1000 search nodes, making it a total of 2000 search nodes. At each subsequent round $i$, each flaw selection strategy is permitted to generate up to $1000 \cdot 2^{i-2}$ additional nodes. The maximum number of nodes generated using a specific flaw selection strategy is $1000 \cdot 2^{i-1}$ after $i$ rounds. An optional upper limit on the number of generated search nodes can be set for each flaw selection strategy. This is useful for flaw selection strategies that typically solve problems quickly, when they solve them at all within reasonable time. Table 7 shows the search limits used by VHPOP at IPC3. These limits were determined after some initial trials on the competition problems. Note that there was no set search limit for the last flaw selection strategy. Whenever the other three strategies all reached their search limits without having found a solution, LCFR-Loc-Conf was used until physical resource limits were reached.

| Name | Order | STRIPS Limit | Durative Limit |
|------|-------|-------------|----------------|
| MW-Loc | 1 | 10000 | 12000 |
| MW-Loc-Conf | 2 | 100000 | 100000 |
| LCFR-Loc | 3 | 200000 | 240000 |
| LCFR-Loc-Conf | 4 | $\infty$ | $\infty$ |

Table 7: Execution order of flaw selection strategies used at IPC3, and also search limits used with each strategy on domains with and without durative actions.

Table 8 shows the number of plans generated in the STRIPS Satellite domain before a solution is found for the four flaw selection strategies used at IPC3, and also the number of generated plans when combining the four strategies using the schedule in Table 7. To better understand how the round-robin scheduling works, we take a closer look at the numbers for problem 15. Table 9 shows how the total number of generated plans is divided between rounds and flaw selection strategies. Note that although MW-Loc is actually the best strategy for this problem, it is stopped already in round 5. The total number of generated plans does not exactly match the actual number of generated plans reported in Table 8. This is because we only consider suspending the use of a flaw selection strategy after all refinements of the last selected plan have been added, so the limit in a round can be exceeded slightly in practice. The numbers in Table 9 represent an idealized situation where flaw selection strategies are switched when the number of generated plans exactly matches the limit for the current round.

VHPOP solved 122 problems out of 224 attempted at IPC3. The quality of the plans, in terms of number of steps, generated by VHPOP was generally very high. For plain STRIPS domains, VHPOP's plans were typically within 10 percent of the best plans found by any planner in the competition, with 28 of VHPOP's 68 STRIPS plans being at least as short as the best plans found and, being a POCL planner, VHPOP automatically exploits parallelism in planning domains, generating plans for STRIPS domains with low total plan execution time (Table 10). Table 11 shows that VHPOP also performed well in terms of number of solved problems in four of the six STRIPS domains, being competitive with top performers such as MIPS and LPG (particularly in the Rovers domain).

| Problem | MW-Loc | MW-Loc-Conf | LCFR-Loc | LCFR-Loc-Conf | All |
|---------|--------|-------------|----------|---------------|-----|
| 1  | 118    | 118         | 118      | 118           | 118   |
| 2  | 229    | 229         | 249      | 249           | 229   |
| 3  | 172    | 172         | 172      | 172           | 172   |
| 4  | 738    | 843         | 822      | 1797          | 738   |
| 5  | 448    | 723000†     | 1018     | 706000†       | 448   |
| 6  | 636000†| 629000†     | 720      | 834           | 2727  |
| 7  | 571    | 745         | 620      | 688000†       | 571   |
| 8  | 482000†| 874         | 1017     | 783           | 1874  |
| 10 | 1245   | 1178        | 1323     | 1275          | 4283  |
| 11 | 1172   | 1172        | 1172     | 1172          | 4172  |
| 12 | 3517   | 3733        | 525000†  | 525000†       | 9542  |
| 13 | 6241   | 382000†     | 559000†  | 544000†       | 18265 |
| 14 | 2352   | 2352        | 2157     | 2157          | 8365  |
| 15 | 74738  | 444000†     | 107375   | 465000†       | 281387|
| 16 | 533000†| 529000†     | 3442     | 3571          | 13471 |
| 17 | 2975   | 2975        | 3438     | 3438          | 8981  |
| 18 | 1584   | 1584        | 1724     | 1724          | 4588  |

Table 8: Number of generated plans in the STRIPS Satellite domain for the four different flaw selection strategies used by VHPOP at IPC3. The rightmost column is the number of plans generated by VHPOP before finding a solution when using the schedule in Table 7. A dagger (†) means that the planner ran out of memory (800 Mb) after generating at least the indicated number of plans.

| Round | MW-Loc | MW-Loc-Conf | LCFR-Loc | LCFR-Loc-Conf | Total |
|-------|--------|-------------|----------|---------------|-------|
| 1 | 1000  | 1000   | 1000  | 1000  | 4000  |
| 2 | 1000  | 1000   | 1000  | 1000  | 4000  |
| 3 | 2000  | 2000   | 2000  | 2000  | 8000  |
| 4 | 4000  | 4000   | 4000  | 4000  | 16000 |
| 5 | *2000* | 8000  | 8000  | 8000  | 26000 |
| 6 | -     | 16000  | 16000 | 16000 | 48000 |
| 7 | -     | 32000  | 32000 | 32000 | 96000 |
| 8 | -     | *36000* | 43375 |       | 79375 |
| Total | 10000 | 100000 | 107375 | 64000 | 281375 |

Table 9: A closer look at the round-robin scheduling for problem 15 in the STRIPS Satellite domain. Italic entries indicate that the search limit for a flaw selection strategy was reached in the round.

| Domain | # Solved | # Steps | # Best | Execution Time | # Best |
|--------|----------|---------|--------|----------------|--------|
| DriverLog | 14 | 1.09 | 5 | 1.15 | 4 |
| ZenoTravel | 13 | 1.04 | 7 | 1.20 | 5 |
| Satellite | 17 | 1.07 | 7 | 1.25 | 5 |
| Rovers | 20 | 1.08 | 7 | 1.08 | 13 |

Table 10: Relative plan quality for the STRIPS domains where VHPOP solved more than half of the problems. There are two plan quality metrics. Number of steps is simply the total number of steps in a plan, while execution time is the total time required to execute a plan (counting parallel actions as one time step). The table shows the average ratio of VHPOP's plan quality and the quality of the best plan generated by any planner, and the number of problems in each domain where VHPOP found the best plan is also shown.

| Planner | Depots | DriverLog | ZenoTravel | Satellite | Rovers | FreeCell | Total |
|---------|--------|-----------|------------|-----------|--------|----------|-------|
| FF | 22 | 15 | 20 | 20 | 20 | 20 | 117 |
| LPG | 21 | 18 | 20 | 20 | 12 | 18 | 109 |
| MIPS | 10 | 15 | 16 | 14 | 12 | 19 | 86 |
| SIMPLANNER | 22 | 11 | 20 | 17 | 9 | 12 | 91 |
| STELLA | 4 | 10 | 18 | 14 | 4 | 0 | 50 |
| VHPOP | 3 | 14 | 13 | 17 | 20 | 1 | 68 |

Table 11: Number of problems solved by top performing fully automated planners in STRIPS domains.

In domains with durative actions[5], total execution time was given as an explicit plan metric, and the objective was to minimize this metric. The specification of an explicit plan metric is a feature of PDDL2.1 not present in earlier versions of PDDL. As VHPOP currently ignores this objective function and always tries to find plans with few steps, it should come as no surprise that the quality of VHPOP's plans for domains with durative actions was significantly worse than the quality of the best plans found (Table 12).[6] VHPOP still produced plans with few steps, however, with over 60 percent of VHPOP's plans for domains with durative actions having the fewest steps. The plan selection heuristic that VHPOP uses is tuned for finding plans with few steps, and it would need to be modified in order to find plans with shorter total execution time. Table 13 shows that LPG solved by far the most problems in domains with durative actions, but that VHPOP was competitive with MIPS and clearly outperformed TP4 and TPSYS.

---

5. There were two types of domains with durative actions at IPC3: "SimpleTime" domains having actions with constant duration and "Time" domains with durations being functions of action parameters. The results with durative actions presented in this paper are for "SimpleTime" domains as there is currently no support for durations as functions of action parameters in VHPOP. It would, in principle, not be hard to add such support though, and we expect future versions of VHPOP to have it.

6. The poor performance is in part also due to the use of 1 for $\epsilon$ (see Section 4) in VHPOP, while most other planners used 0.01 or less. Using 0.01 for $\epsilon$ with VHPOP reduces the total execution time of plans with about 15 percent.

| Domain | # Solved | # Steps | # Best | Execution Time | # Best |
|--------|----------|---------|--------|----------------|--------|
| DriverLog | 14 | 1.04 | 8 | 1.50 | 0 |
| ZenoTravel | 13 | 1.04 | 10 | 1.54 | 0 |
| Satellite | 17 | 1.04 | 9 | 2.51 | 0 |
| Rovers | 7 | 1.04 | 5 | 1.39 | 0 |

Table 12: Same information as in Table 10, but for domains with durative actions.

| Planner | Depots | DriverLog | ZenoTravel | Satellite | Rovers | Total |
|---------|--------|-----------|------------|-----------|--------|-------|
| LPG | 20 | 20 | 20 | 20 | 12 | 92 |
| MIPS | 11 | 15 | 14 | 9 | 9 | 58 |
| TP4 | 1 | 2 | 5 | 3 | 4 | 15 |
| TPSYS | 0 | 2 | 2 | 2 | 4 | 10 |
| VHPOP | 3 | 14 | 13 | 17 | 7 | 54 |

Table 13: Number of problems solved by fully automated planners in domains with durative actions.

While VHPOP was a top performer at IPC3 in terms of plan quality, it was far from the top in terms of planning time. VHPOP was typically orders of magnitude slower than the fastest planner. The high planning times for VHPOP can in part be attributed to implementation details. Improvements to the code (e.g. using pointer comparison instead of string comparison whenever possible) since the planning competition has resulted in 10 to 20 percent lower planning times when using ground actions and when using lifted actions the planner is more than twice as fast as before. The reachability analysis is still a bottleneck, however, and further improvements could definitely be made there. It is important to remember, though, that we basically run four planners at once by using four flaw selection strategies concurrently. Table 14 shows the average relative performance of VHPOP at IPC3 compared to the performance of VHPOP using only the best flaw selection strategy for each problem. VHPOP with the best flaw selection strategy is on average two to three times faster than VHPOP with four concurrent strategies. Using several flaw selection strategies simultaneously helps us solve more problems, but the price is reduced speed. By more intelligently scheduling the different flaw selection strategies depending on domain and problem features, and not just using a fixed schedule for all problems, we could potentially increase planner efficiency significantly.

## 6. Discussion

McDermott (2000) finds the absence of POCL planners at the first planning competition in 1998 striking, as such planners had been dominating planning research just a few years earlier. "It seems doubtful that the arguments in [POCL planners'] favor were all wrong, and it would be interesting to see partial-order planners compete in future competitions", McDermott writes. After two competitions without POCL planners, we believe that VHPOP's performance at IPC3 demonstrates that POCL planning—at least with ground actions—can be competitive with CSP-based and heuristic state space planning. VHPOP also shows that temporal POCL planning can be made practical by using the same heuristic techniques that have been developed for classical planning. The idea of using the POCL paradigm for temporal planning is not new and goes back at least to Vere's DEVISER (Vere, 1983), but we are the first to demonstrate the effectiveness of temporal POCL planning on a larger set of benchmark problems.

| Domain | STRIPS | Durative |
|--------|--------|----------|
| DriverLog | 2.52 | 2.66 |
| ZenoTravel | 2.76 | 2.86 |
| Satellite | 1.78 | 2.01 |
| Rovers | 2.32 | 3.37 |

Table 14: Each number in the table represents the average ratio of the planning time for VHPOP using all four flaw selection strategies concurrently and the planning time for VHPOP with only the fastest flaw selection strategy.

We hope that the success of VHPOP at IPC3 will inspire a renewed interest in plan space planning, and we have made the source code for VHPOP, written in C++, available to the research community in an online appendix so that others can build on our effort.[7]

While VHPOP performed well above our expectations at IPC3, we see several ways in which we can further improve the planner. Speed, as mentioned in Section 5, is the principal weakness of VHPOP. The code for the reachability analysis is not satisfactory, as it currently generates ground action instances before performing any reachability analysis. This often leads to many ground action instances being generated that do not have preconditions with finite heuristic cost (according to the additive heuristic). We believe that VHPOP could profit from code for reachability analysis in well-established planning systems such as FF. We could also improve speed by better scheduling different flaw selection strategies. We would like to see statistical studies, similar to that of Howe et al. (1999), linking domain and problem features to the performance of various flaw selection strategies.

We have so far only considered using different flaw selection strategies. However, running multiple instances of VHPOP using different plan selection heuristics could be equally interesting. We have, for example, noticed that using the additive heuristic *without* accounting for reuse helps us solve two more problems in the Satellite domain. It would also be interesting to have the FF heuristic implemented in VHPOP and see how well it performs in a plan space planner, possibly using local search techniques instead of $A^*$. It is not likely, however, that the results on local search topology for the FF heuristic in state space (Hoffmann, 2001) carry over to plan space. While many of the benchmark planning domains contain actions whose effects can be undone by other actions, the plan operators causing transitions in the search space of a plan space planner are different from the actions defined for a planning domain, and the effects of a plan space operator are generally irreversible. We would likely need to add *transformational* plan operators that can undo linking and ordering decisions. Incidentally, VHPOP started out as a project for adding transformational plan operators to UCPOP, but we got side-tracked by the need for better search control, and our research on transformational POCL planning was suspended. With the recent improvements in search control for POCL planners, it may be worthwhile to once again consider adding transformational plan operators.

---

7. The latest version of VHPOP is available for download at www.cs.cmu.edu/˜lorens/vhpop.html.

In addition to considering different search control heuristics, we could also have instances of VHPOP working with lifted actions instead of ground actions. Recent improvements to the code have significantly reduced the overhead for maintaining binding constraints, making planning with lifted actions look considerably more favorable than was reported in earlier work (Younes & Simmons, 2002). Planning with lifted actions could be beneficial for problems with a high branching factor in the search space due to a large number of objects.

We would also like to see support for numeric effects and preconditions in future versions of VHPOP. This would make VHPOP fully compatible with PDDL2.1. We have also mentioned the need for plan ranking heuristics better tailored for temporal planning, so that VHPOP's performance in terms of plan execution time for domains with durative actions can approach the performance of the best temporal planners at IPC3.

## Acknowledgments

This effort was sponsored by the Defense Advanced Research Project Agency (DARPA) and the Army Research Office (ARO), under contract no. DAAD19-01-1-0485. The U.S. Government is authorized to reproduce and distribute reprints for Governmental purposes notwithstanding any copyright annotation thereon. The views and conclusions contained herein are those of the authors and should not be interpreted as necessarily representing the official policies or endorsements, either expressed or implied, of DARPA, ARO, or the U.S. Government.